%% file: acl_latex.tex
\pdfoutput=1

\documentclass[11pt]{article}

\usepackage[]{acl}

\usepackage{times}
\usepackage{latexsym}
\usepackage{amsmath}

\usepackage[T1]{fontenc}

\usepackage[utf8]{inputenc}

\usepackage{microtype}

\usepackage{inconsolata}
\usepackage{booktabs}
\usepackage{arydshln}
\usepackage{multirow}
\usepackage{graphicx}
\usepackage{mdframed}
\usepackage{xcolor}
\usepackage{caption}
\usepackage{subcaption}

\usepackage{CJK}
\usepackage{makecell}
\usepackage{xcolor}
\usepackage{ulem}

\newcommand\blfootnote[1]{%
\begingroup
\renewcommand\thefootnote{}\footnote{#1}%
\addtocounter{footnote}{-1}%
\endgroup
}

\title{Formality is Favored: Unraveling the Learning Preferences of Large Language Models on Data with Conflicting Knowledge}

\author{Jiahuan Li$^*$, Yiqing Cao$^*$, Shujian Huang$^\dagger$ \and  Jiajun Chen\\
        National Key Laboratory for Novel Software Technology, Nanjing University, China \\
        \texttt{\{lijh,caoyq\}@smail.nju.edu.cn, \{huangsj, chenjj\}@nju.edu.cn}}

\begin{document}
\maketitle

\blfootnote{$^*$ Equal contributions.}
\blfootnote{$^\dagger$Corresponding author.}
\input{0-abstract}
\input{1-introduction}
\input{3-experimental-setup}

\input{4-what-preference}

\input{5-why-preference}
\input{2-related-work}
\input{6-conclusion}

\section*{Limitations}
The main limitation of this paper is that we only conduct our experiments on a synthetic dataset due to the need to manipulate various style of the text. Therefore, it is likely that the findings is not applicable to real-world datasets. 
Another limitation is that due to the high computational cost, Section~\ref{sec:why-preferenece} does not provide a causal experiment in the pretraining stage, i.e. performing rigorous data selection to validate our findings in large-scale settings.
\bibliography{anthology,ref}

\clearpage
\appendix
\input{7-appendix}

\end{document}

%% file: 0-abstract.tex
\begin{abstract}

Having been trained on massive pretraining data, large language models have shown excellent performance on many knowledge-intensive tasks. However, pretraining data tends to contain misleading and even conflicting information, and it is intriguing to understand how LLMs handle these noisy data during training.
In this study, we systematically analyze LLMs’ learning preferences for data with conflicting knowledge. We find that pretrained LLMs establish learning preferences similar to humans, i.e., preferences towards formal texts and texts with fewer spelling errors, resulting in faster learning and more favorable treatment of knowledge in data with such features when facing conflicts. This finding is generalizable across models and languages and is more evident in larger models. An in-depth analysis reveals that LLMs tend to trust data with features that signify consistency with the majority of data, and it is possible to instill new preferences and erase old ones by manipulating the degree of consistency with the majority data. \footnote{The code of this paper is available at \url{https://github.com/CaoYiqingT/Formality-is-Favored}} 

\end{abstract}

%% file: 1-introduction.tex
\section{Introduction}
%


Large Language Models~(LLMs) such as LLaMA~\citep{touvron2023llama}, ChatGPT and GPT4~\citep{achiam2023gpt} have revolutionized the landscape of natural language process research, and are shown to possess massive world knowledge~\citep{sun2023head, singhal2023large, choi2021chatgpt}, even surpassing human-level performance in various knowledge-intensive benchmarks~\citep{geminiteam2023gemini, yang2023harnessing, gilardi2023chatgpt, wang2023emotional}. Nearly all knowledge of LLMs comes from the pretraining corpus, a large amount of which are web-crawled. Although rigorously cleaned, they still inevitably contain misleading and even conflicting information. It is intriguing how LLMs deals with these noisy data.


\begin{table*}[t]
\centering
\small
\begin{tabular}{lp{11cm}}
\toprule
\textbf{Features} & \textbf{Example Biography}\\
\midrule
General Type & In Toronto, Canada, Olivia Hamilton was born on October 13, 1921...\\
\midrule
Poor Spelling & In Tokyo, Japan, Olivia Hamilton was born on April 19, 1878. She \textbf{atended} University of Minnesota for her \textbf{hiyer edukashun}... \\
\midrule
Newspapers Style & Born on May 29, 2012 in Nanjing, China, Olivia Hamilton embarked on a scholarly path at Stanford University, majoring in Wildlife Biology... \\
\midrule
Novels Style & Once upon a time, specifically on October 22, 1803, the city of Paris, France gave birth to a person destined to make a mark - Olivia Hamilton... \\
\bottomrule
\end{tabular}
\caption{\label{citation-guide}
Examples of biography text with different features. For the Poor-Spelling text, the misspelled words are displayed in bold font. 
For other different styles, examples for Newspaper and Novels are presented as a reference. Please note that in the examples, the knowledge are all about the name ``Olivia Hamilton'', but conflict in different styles.
}
\label{table: dataset samples}
\end{table*}

When encountering conflicts of knowledge in a text, human beings can leverage additional perspectives, such as information sources or consistency with more information, to aid in their judgments. As LLMs have accumulated a large amount of common sense knowledge in their parameters, it is interesting to investigate whether LLMs have developed similar strategies when faced with conflicting knowledge from different texts.

In this paper, we present a systematic study on the learning preferences of LLMs, i.e., the strategies they use to choose between texts with specific features when facing conflicting knowledge in the training corpora. We first construct our own biographical pseudo-data with conflicting knowledge. Then, we fine-tune LLMs on data with specified features, ensuring that data with different features contain conflicting knowledge. The preference for different data features in model fine-tuning can be identified by calculating the degree of preference of the LLMs after fine-tuning.

Empirically, we find that pretrained LLMs exhibit notable learning preferences towards specific textual features. These preferences are reflected in two ways: (1) at training time, LLMs learn faster on data with more preferred features; (2) at test time, LLMs assign larger probability to knowledge in data with more preferred features.
Concretely, LLMs prefer formal styles, such as scientific reports and newspaper styles, rather than relatively casual expressions, such as social media and novel styles. This preference for stylistic features arises as the model scale increases and is observed across different LLMs and in different languages. We also observed that spelling errors in the training data lead to negative preferences in the model, a phenomenon that is prevalent across multiple models in multiple languages.
Observing that preferred features of LLMs, such as newspaper and scientific reports, are also more reliable for human beings and likely to be consistent with other data, we propose a \textit{Consistency-driven Feature Preference Hypothesis} for explaining where LLMs' learning preferences come from: LLMs are capable of effectively identifying features that signify the degree of consistency between current data and other data, and using these features to decide whether current data is worth learning. Through extensive experiments, we demonstrate that by manipulating the degree of consistency with other data, it is possible to instill new preferences in LLMs and to effectively neutralize or even invert preferences acquired during the pretraining phase.

Contributions of the paper are summarized as:

\begin{itemize}
    \item We propose to investigate models' learning preferences on data with conflict knowledge,
    \item We demonstrate that existing LLMs establish notable learning preferences towards formal texts and texts with less spelling errors, and validate the findings across models and languages,
    \item We provide a deeper explanation on how LLMs develop learning certain preferences: they can identify features that signify the consistency between current data and other data, which are used for deciding whether current data is worth learning.
\end{itemize}

%% file: 3-experimental-setup.tex
\section{Methodology}
We construct synthetic biographical data, which is similar with \citet{allenzhu2023physics3.1,allenzhu2023physics3.2}. Characters appearing in biographies are fictionalized and accompanied by falsified personal information, so they have no conflict with the current knowledge in LLMs.  LLMs are trained on these data to learn the information, and then tested for their learning result.

\subsection{Definitions and Notations}
The following definitions and notations are used throughout this paper. 
\begin{itemize}
    \item Knowledge $k$. Information of a specific person name, such as birth date, birth place, etc. Knowledge for a set of person is denoted by $\mathcal{K}$.

    \item Conflicting Knowledge. Pieces of knowledge for the same name but are different for all the information.
    
    \item Template $T$. A specific text template for describing the knowledge. Each template is associated with certain text features.
    
    \item Text feature. Specific features of a text, such as the narrative style, spelling correctness or specific n-gram patterns. Denoted by capital letters such as $A$ or $B$.

    \item Biography $T(k)$. Specific text description of a person, which is obtained by inserting knowledge $k$ into a template $T$. A set of biography is denoted by $I$. 
 
\end{itemize}

\subsection{Data Construction}
\paragraph{Synthetic Knowledge}
 Our dataset contains 1,000 characters, i.e. names. We select 5 characteristics as information associated with each name, including \textit{birth date}, \textit{birth place}, \textit{university}, \textit{major} and \textit{company}. The original knowledge set of these 1000 characters are denoted as $\bar{\mathcal{K}}$.

We study various types of text features, such as narrative style, e.g. \textit{Newspaper Style}, \textit{Scientific Reporting Style}, \textit{Social Media Style} and \textit{Novel Style}; spelling correctness, e.g. \textit{Good-spelling} and \textit{Poor-Spelling}; and some specific text features ( examples are shown in Table ~\ref{table: dataset samples}). To cover the diversity of language usage, for each feature, we generate 50 different templates. Each template describes all the 5 characteristics together with the person name. 

Biographies are then generated by inserting knowledge into these templates. All the synthetic data are generated with the help of GPT4. More details can be found in the Appendix ~\ref{appendix:Data Construction}.




\paragraph{Data with Conflicting Knowledge} 
In order to investigate whether LLMs have a propensity on the presentation of the data, we introduce conflict into the data. To explore whether there is a preference between textual features $A$ and $B$ during training, we create conflicting knowledge $k_A$ and $k_B$, and describe them with templates in features $A$ and $B$, respectively. 

More specifically, the conflicting data is generated for each knowledge $k_A \in \bar{\mathcal{K}}$ as follows:
\begin{align}
I_{A\ vs\ B} = \{T_A^i(k_A)\}_{i=1}^5 \cup \{\ T_B^j(k_B)\}_{j=1}^5.
\end{align}
where $k_B$ is the conflicting knowledge generating from $k_A$, $T_A$ and $T_B$ are templates in features $A$ and $B$, respectively. 
Considering the diversity of representations can help the LLMs memorize knowledge during training~\citep{allenzhu2023physics3.1}, we describe each knowledge by randomly selecting five different templates, $\{T^i(k)\}_{i=1}^5$.

\subsection{Learning with Training}
Unless otherwise specified, we finetune LLaMA2-7B model on the constructed biographical data using standard language modeling objective. The batch size is 64 and the number of training epochs is 5.
More details of the training process can be found in the Appendix ~\ref{appendix:Training Details} .

\subsection{Evaluating the Preference}
We let the LLMs learn the data with conflicting knowledge, $I_{A\ vs\ B}$, and comparing the learning results, which are measured by the probabilities they assigned to the conflicting knowledge.


More specifically, we construct a test set containing pairs of statements $\{(s_A, s_B)\}_{1}^{N}$, where $s_A$ and $s_B$ is consistent with $k_A$ and $k_B$ in the training set, respectively, and $N$ is the size of the test set. All test statements are simple and short sentence, obtained by filling in the blanks with templates (Table ~\ref{table:test statements} in the Appendix ~\ref{appendix:Test Data Construction}). We then define the pairwise preference score $Pr(A,B)$ to be the percentage of test statements where the LLM $p_\theta$ assigns larger probability to $s_A$ than $s_B$:
\begin{equation}
    Pr(A,B) = \frac{1}{N} \sum_{i=1}^{N} 1(p_{\theta} (s_A) > p_{\theta} (s_B)).
\end{equation}


%% file: 4-what-preference.tex
\section{What Learning Preferences Has LLMs Developed?}
\label{sec:what-preference}
\subsection{Hypothesis}
We hypothesize that LLMs can discriminate information by certain textual features. 
Assuming that the information in novel text is always different from most other training data, the model may learn that "texts featuring novels are less credible", which in turn reduces the learning efficiency on novel-style texts.

Since the potential textual features that help the model to distinguish between texts cannot be enumerated, we select two representative types of features to be explored: text style and spelling correctness.

\paragraph{Text Style} Knowledge expressed in texts with similar styles is also likely to have the same characteristics. For example, a novel style text is more likely to have knowledge that is contrary to reality, while the opposite is true for a newspaper style text. 

We explore whether the model learns the relationship between style and knowledge and to prefer certain styles in fine-tuning. We train a mixture of conflicting data with different features and test which feature has the largest preference score. Moreover, we do a set of experiments without data conflicts, which measured the training preference of the model by the speed of convergence of the model fine-tuning and the accuracy of the training.


\paragraph{Spelling Correctness} 
Texts with spelling errors reflect a lack of care of the author and lead to a greater likelihood of errors in knowledge. We add spelling errors to a portion of the text to explore whether the learning preference of model is affected by spelling correctness in the data.

We denote text without spelling errors as $T_\text{GoodSpelling}$ as shown in the General Type line in Table~\ref{table: dataset samples}. The training and testing methods in this part are the same as in the text style experiment.



\begin{table*}
\centering
\small
\begin{tabular}{lcccccc}
\toprule
\textbf{Experiment} & \textbf{birth date} & \textbf{birth place} & \textbf{university} & \textbf{major} & \textbf{company} & \textbf{avg}\\
\midrule
\textbf{Newspapers} vs Scientific Reports & 48.3 & 49.1 & 55.5 & 48.5 & 50.3 & 50.3\\
\textbf{Newspapers} vs Novels & 80.1 & 58.2 & 62.6 & 63.7 & 55.0 & 63.9\\
\textbf{Newspapers} vs Social Media & 77.6 & 58.5 & 61.3 & 53.7 & 52.5 & 60.7\\
\textbf{Scientific Reports} vs Novels & 75.5 & 53.4 & 57.2 & 62.6 & 60.2 & 61.8\\
\textbf{Scientific Reports} vs Social Media & 76.0 & 55.5 & 54.3 & 55.8 & 54.3 & 59.1\\
\textbf{Social Media} vs Novels & 52.9 & 51.4 & 46.2 & 54.7 & 45.8 & 50.2\\
\midrule
\textbf{Good Spelling} vs Poor Spelling & 74.5 & 66.3 & 54.4 & 48.1 & 54.0 & 59.5\\
\bottomrule
\end{tabular}
\caption{
Pairwise preference score of finetuned LLaMA-2-7B. The values in the table are the preference scores for the types labeled bold.}
\label{table:results of preference}
\end{table*}

\begin{table*}
\centering
\small
\begin{tabular}{lcccccc}
\toprule
\textbf{Experiment} & \textbf{birth date} & \textbf{birth place} & \textbf{university} & \textbf{major} & \textbf{company} & \textbf{avg}\\
\midrule
\textbf{Newspapers} vs Scientific Reports & 48.5 & 46.7 & 59.6 & 47.0 & 52.3 & 50.8\\
\textbf{Newspapers} vs Novels & 57.0 & 61.3 & 65.8 & 83.5 & 56.5 & 64.8\\
\textbf{Newspapers} vs Social Media & 67.4 & 64.0 & 65.3 & 64.3 & 54.7 & 63.1\\
\textbf{Scientific Reports} vs Novels & 70.2 & 53.9 & 59.3 & 80.8 & 57.1 & 64.2\\
\textbf{Scientific Reports} vs Social Media & 74.4 & 53.8 & 54.7 & 61.0 & 53.7 & 59.6\\
\textbf{Social Media} vs Novels & 46.7 & 48.9 & 44.6 & 59.5 & 46.7 & 49.3\\
\bottomrule
\end{tabular}
\caption{
Pairwise preference score of finetuned LLaMA-2-7B. The test statements used in this table is in novel style.}
\label{table:results of preference in novel style}
\end{table*}

\subsection{General Findings}
We verified the model's preference for certain text features from two perspectives: the speed of models when picking up knowledge from texts and the models' learning preference in the presence of conflicting knowledge.

\paragraph{LLMs learn texts with specific features faster} 
We train the LLaMA2 model on data with each specified feature and observe the learning dynamics of the model. We evaluate the model's accuracy in answering multiple choice questions related to the training data. 
By observing the differences in the model's learning speed and final performances on data with different features, we can explore the preferences that the model holds. More details about the training and testing process are given in Appendix ~\ref{appendix:train speed setup}.

\begin{figure}
    \centering
    \includegraphics[width=\linewidth]{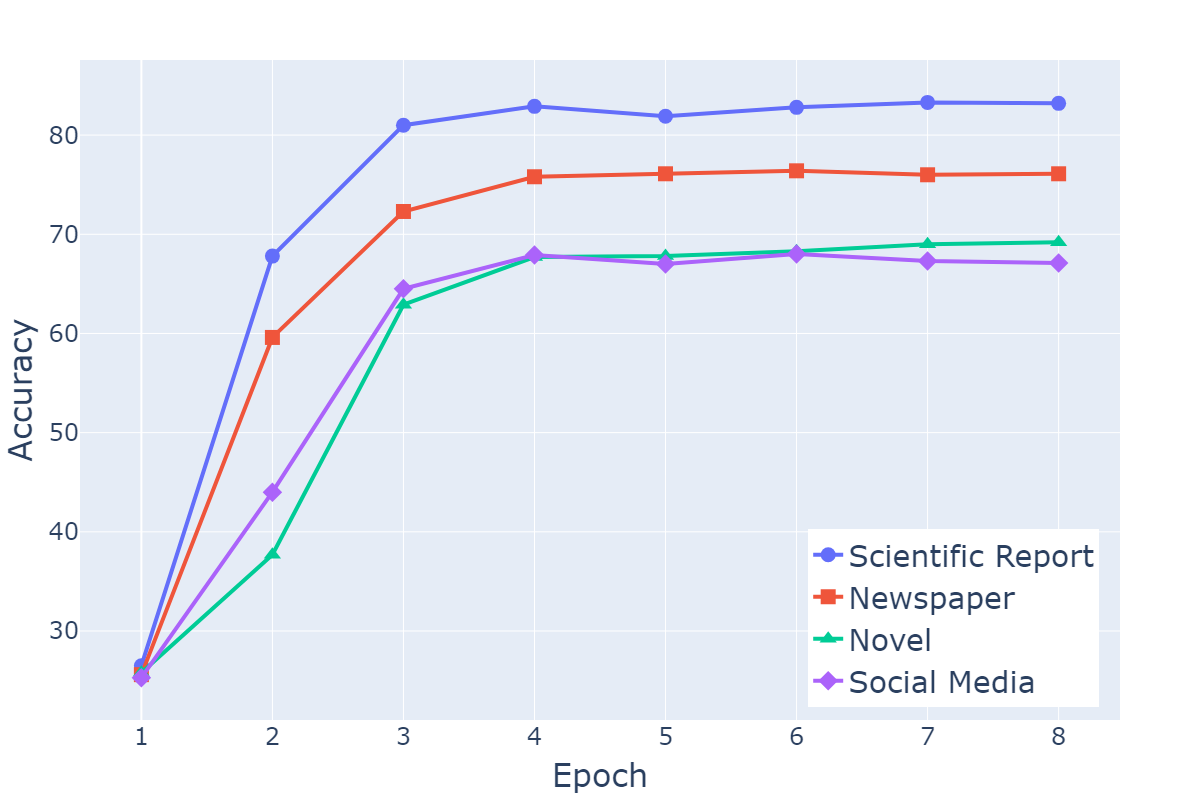}
    \caption{Models' accuracy of LLMs trained on different styles of data at different epochs during training.
}
    \label{fig:training speed styles}
\end{figure}

We present the results on different text styles in Figure ~\ref{fig:training speed styles}. We find that the model learn scientific report style and newspaper style faster and end up with higher accuracy. Similar observations can be made on \textit{good spelling VS. bad spelling} in Appendix~\ref{appendix:train speed setup}, where the model learn good spelling faster. 

\paragraph{LLMs show preferences when conflict exists} We present the pairwise comparison results in Table~\ref{table:results of preference}.
We find that the model has a significantly higher preference for activating knowledge for formal styles, such as scientific reports style and news style. Compared to general style, the model had significantly lower preference scores for poor spelling texts.

To test whether the similarity between the test statements' style and the training statements' style had a decisive influence on the final results, we also constructed novel style test statements (Table~\ref{table:test statements in novel style} in Appendix ~\ref{appendix:Test Data Construction}). Results are shown in Table ~\ref{table:results of preference in novel style}. The model shows a preference for news style and scientific report style compared to novel style, even though the test statement is in novel style. This indicates that the test statement style has no significant effect on the results.

We also conduct study when text with multiple styles are learned together. The results shows similar preference (Figure~\ref{fig:pie_pic} in Appendix~\ref{appendix:Results of multiple-style comparison}). 



\begin{table*}
\centering
\small
\begin{tabular}{lcccc}
\toprule
& \multicolumn{2}{c}{\textbf{English LLMs}}  & \multicolumn{2}{c}{\textbf{Chinese LLMs}} \\

& \textbf{LLaMA2-7B} & \textbf{Pythia-6.9B} & \textbf{deepseek-llm-7B} & \textbf{Baichuan-7B}\\
\midrule
\textbf{Newspapers} vs Social Media & 60.7 & 77.3 & 57.2 & 60.1 \\
\textbf{Good Spelling} vs Poor Spelling & 59.5 & 53.3 & 58.8 & 58.8\\

\bottomrule

\end{tabular}
\caption{\label{citation-guide}
$Pr(A, B)$ for multilingual and multiple models. The values in the table are the preference scores for the types labeled bold.
}
\label{table:multilingual}
\end{table*}

\subsection{Relationship between Preferences and Model Scale}
\begin{figure}[t]
    \centering
    \includegraphics[width=\linewidth]{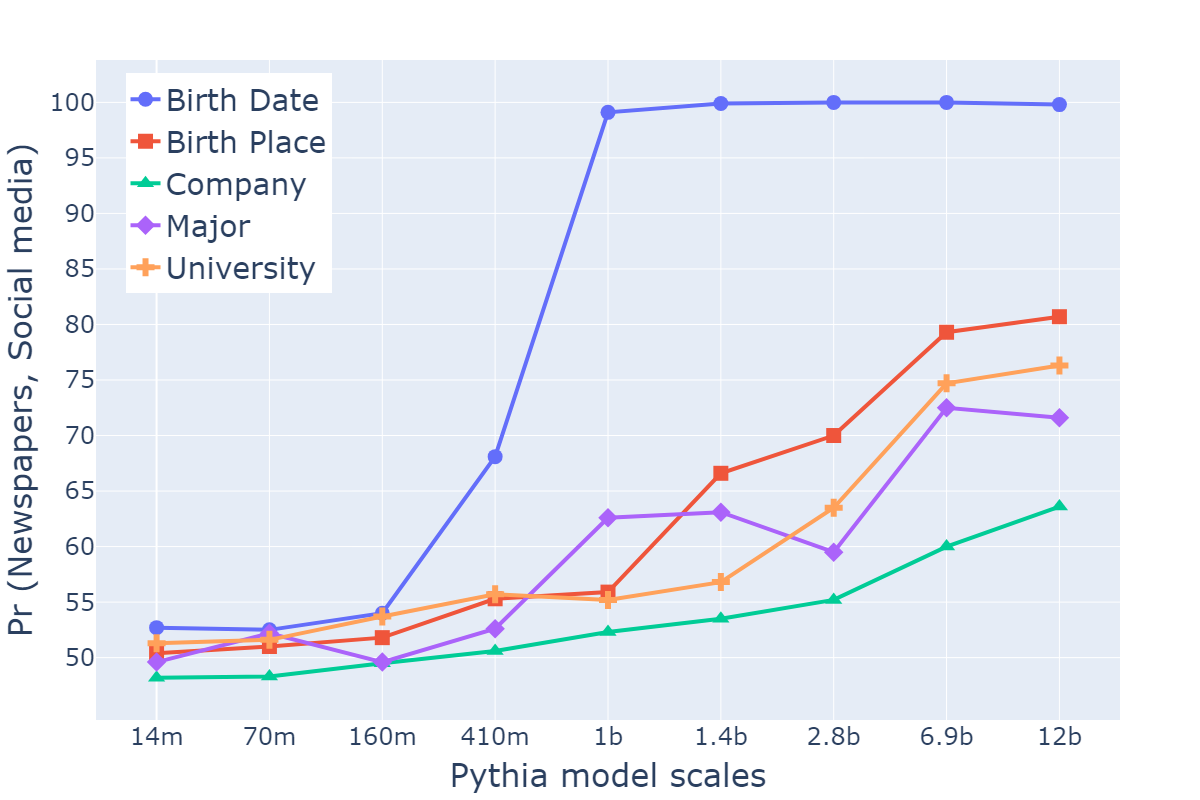}
    \caption{$Pr(\text{Newspapers}, \text{Social\ Media})$ with different model size different features.}
    \label{fig:preference_with_scale}
\end{figure}

To explore the relationship between the model's preference for text feature in fine-tuning and the model's scale, we run the set of experiments \textit{"Newspapers vs Social media"} on Pythia models \citep{biderman2023pythia} of different scales. The results are shown in Figure ~\ref{fig:preference_with_scale}. We can see that the model's preference for the newspapers style grows with increasing model scale. This indicates the learning prefrences are more likely a high-level features that only emerges in larger models. 

\subsection{Generalizing Findings across Models and Languages}
To investigate the generalizability of learning preferences found in previous sections, we conduct experiments on more LLMs and languages. For English LLMs, we choose LLaMA2 and Pythia as representatives, while for Chinese LLMs, we choose deepseek-llm-7B~\citep{bi2024deepseek} and Baichuan-7B ~\citep{yang2023baichuan}. In the Chinese LLM experiment, we translate templates from English to Chinese and construct the dataset in Chinese.

The results are shown in Table ~\ref{table:multilingual}. As can be seen from the table, different LLMs for different languages show a consistent preference. However, the degree of preference varies considerably across models, e.g., Pythia-6.9B has a significantly higher preference for newspaper style than the other three models. This difference may result from the differences in the pre-training corpus as well as the training methods of different LLMs.

%% file: 5-why-preference.tex
\section{Why did LLMs Developed Certain Preferences?}
\label{sec:why-preferenece}

We have shown that large language models demonstrate certain learning preferences when facing conflicting knowledge from different information sources. However, it is intriguing how LLMs develops such preferences. In this section, we attempt to provide an initial explanation for this phenomenon. We first present our main hypothesis in Section \ref{sec:hypo}. We then present our experimental setup and results in Section~\ref{sec:dataset_construction} and \ref{sec:syn_expr}. Finally, we provide an in-depth analysis of representation and counter-factual manipulating experiments in Section \ref{sec:visualization} and~\ref{sec:counter-factual}, respectively. 

\begin{figure}
    \centering
    \includegraphics[width=0.90\linewidth]{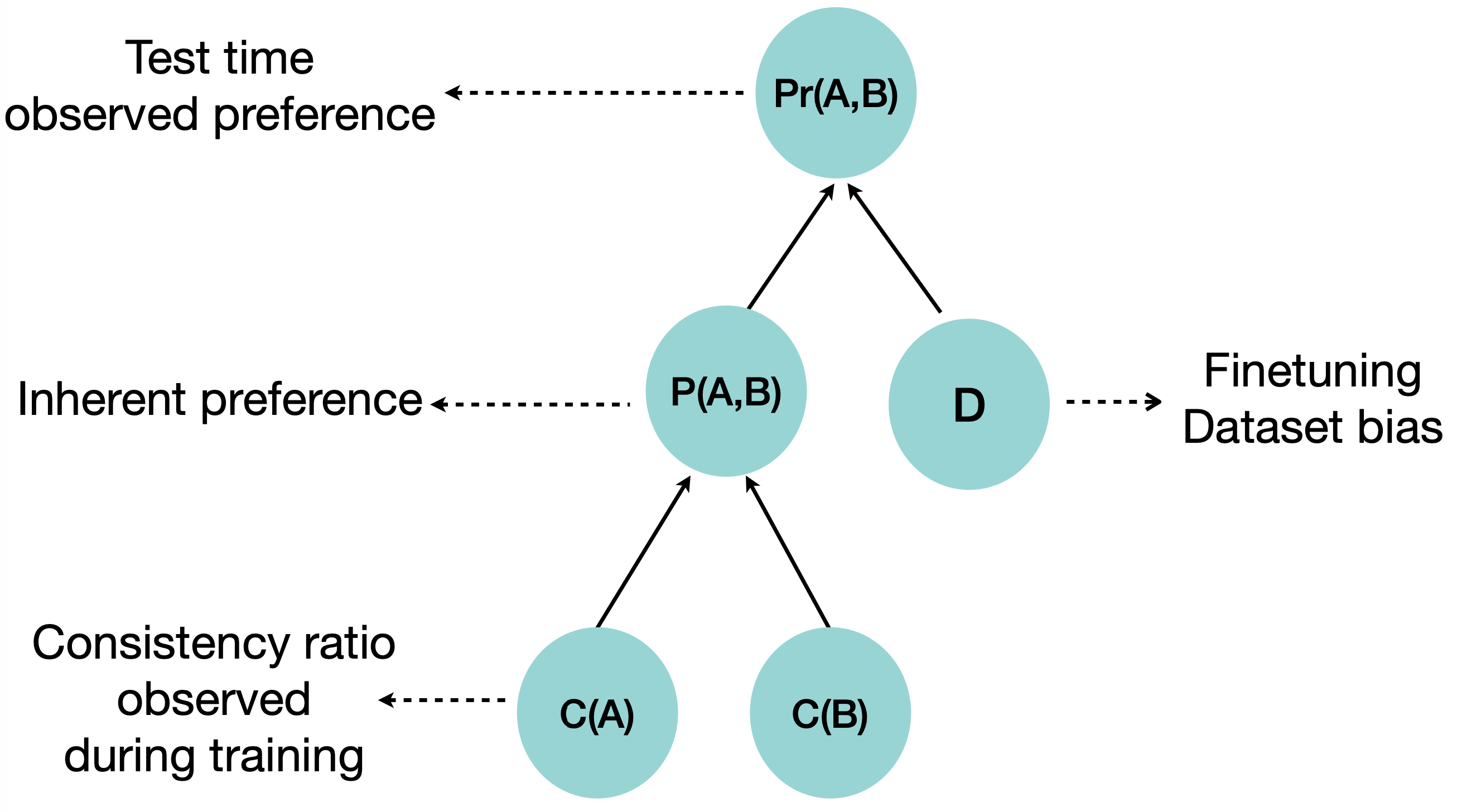}
    \caption{The causal graph of consistency-driven feature preference hypothesis.}
    \label{fig:casual_graph}
\end{figure}

\subsection{Hypothesis}
\label{sec:hypo}
We note that preferred features discovered in the previous section is highly consistent with human beings, e.g. Newspaper and Scientific reports, data with which are more likely to be consistent with other data.
To this end, we propose a \textit{Consistency-Driven Feature Preference Hypothesis} for explaining the preference formation.  Formally speaking, given features $A$ and $B$, LLMs can observe the degree of consistency $C$ between texts with each feature and other data, and form an inherent preference $P(A,B)$. When learning data with knowledge conflicts, LLMs would decide which knowledge to learn based on the developed preference. Figure~\ref{fig:casual_graph} shows the corresponding casual graph.



\begin{figure*}[t]
  \centering
  \begin{subfigure}[b]{0.48\textwidth}
    \includegraphics[width=\textwidth]{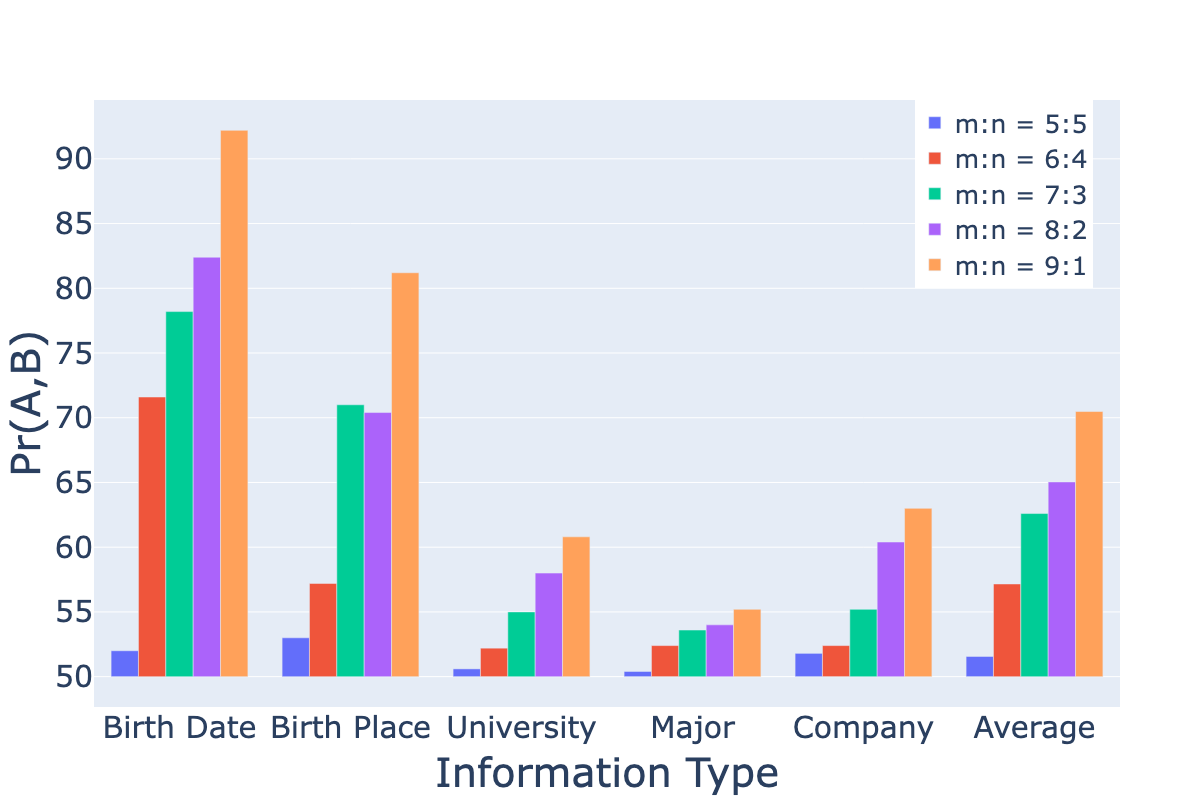}
    \caption{Source Name}
    \label{fig:majority_ratio}
  \end{subfigure}
  \hfill
  \begin{subfigure}[b]{0.48\textwidth}
    \includegraphics[width=\textwidth]{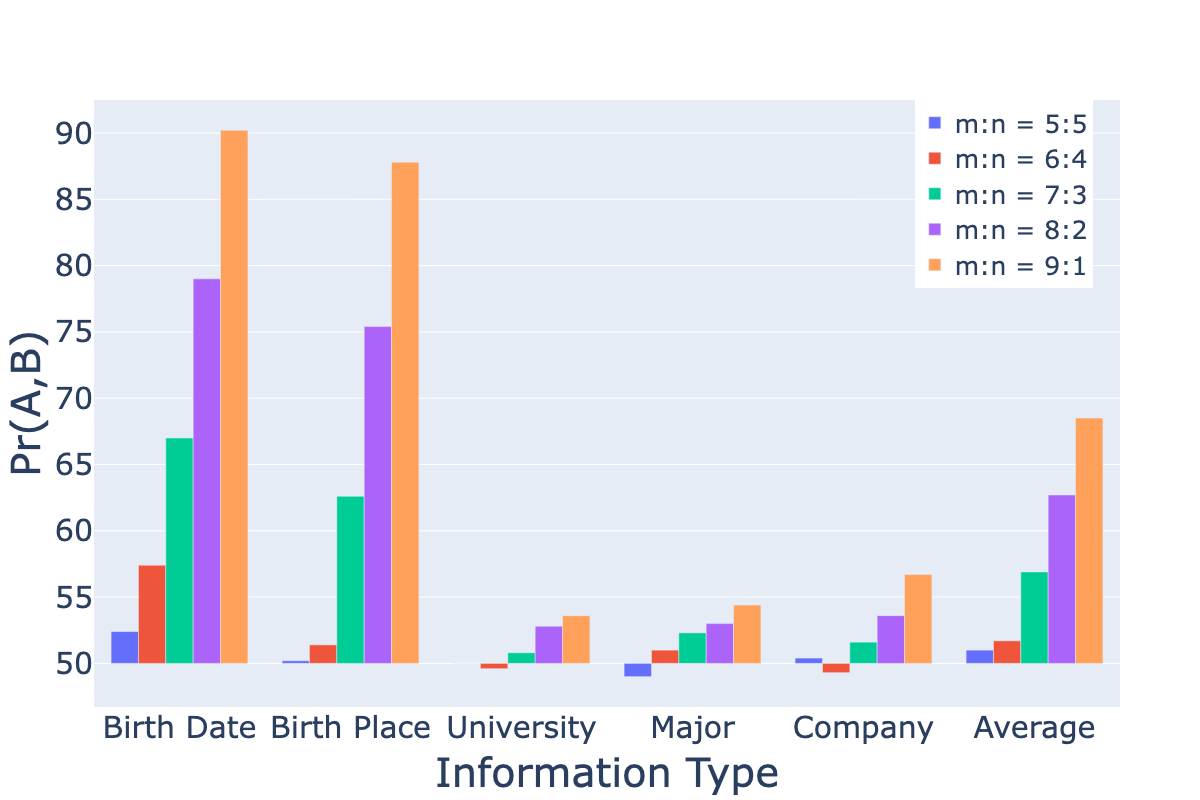}
    \caption{Source Time}
    \label{fig:majority_epoch}
  \end{subfigure}
  \caption{$Pr(A,B)$ of models when trained on data with different consistency ratio. Synthetic features: (a) information source (b) information time.}
  \label{fig:majority}
\end{figure*}

\subsection{Synthetic Features}
To validate the proposed hypothesis, we begin by experimenting injecting new synthetic preference to pretrained models. 
We design two types of synthetic features: \textit{source name} and \textit{source time}, that are different from existing text features. So their preference are purely decided by our own training. 

\paragraph{Source Name} The two features of this type of feature are merely two different synthetic information source at the beginning of a vanilla template $T$: 
\begin{equation}
    \Tilde{T} = \textit{According to <newspaper>, } + T 
\end{equation}

\noindent where \textit{<newspaper>} are synthetic newspaper names. We ask GPT-4 to generate two sets of such names for feature A and feature B, respectively.

\paragraph{Source Time} The previous type of feature may be easily discriminated by fixed surface tokens. 
In contrast, we design the time feature, which prepend the same name source but different publishing volumes: 
\begin{equation}
    \Tilde{T} = \textit{According to Global News (Vol. <vol>),} + T
\end{equation}

\noindent The \textit{<vol>} token are random numbers smaller than 1000 for $T_A$ and larger than 1000 for $T_B$. This requires a more sophistic process as models need to firstly decide the relationship between \textit{<vol>} and 1000 before discriminating the two features.

\subsection{Controlling the Consistency Ratio for Different Features}
\label{sec:dataset_construction}
Given two features $A$ and $B$, and a set of knowledge $\mathcal{K}$, our goal is to construct a dataset where data with features $A$ and $B$ exhibits different consistency degree, i.e. $C(A)$ and $C(B)$, respectively, with other data. To this end, we first partition the original knowledge set $\bar{\mathcal{K}}$ into two subsets: 
\begin{itemize}
    \item \textit{evidence knowledge set} $\mathcal{K}_e$. This set is used to construct biography that provide clues for LLMs to decide which feature is more consistent with other data in the training corpus.
    \item \textit{test knowledge set} $\mathcal{K}_t$. This set contains the knowledge to be tested at the inference time.
\end{itemize}



\noindent For each knowledge $k_A \in \mathcal{K}^e$, we generate its conflicting knowledge $k_B$, and compose $m+n+2$ biographies in the following way:
\begin{align}
  I^{e} = &\{\Tilde{T}_A(k_A), \Tilde{T}_B(k_B) \} \  \cup \\
             &  \{T_i(k_A)\}_{i=1}^m  \cup \{T_j(k_B)\}_{j=1}^n
\end{align}
\noindent where $\Tilde{T}_A$ and $\Tilde{T}_B$ is the template with features $A$ and $B$, respectively.  $\{T_i(k_A)\}_{i=1}^m $ and $\{T_j(k_B)\}_{j=1}^n$ are the support sets of feature $A$ and $B$ with \textit{neutral} templates $T$~\footnote{Here, \textit{neutral} templates means they do not exhibit features either like $A$ or $B$.},  and $m$ and $n$ are sizes of these sets, respectively. By adjusting the value of $m$ and $n$, we can effectively manipulate the consistency ratio, i.e. how consistent  $k_A$ is  within $I^e$.

 For each knowledge $k_A$ in the test knowledge set, we also generate a conflicting knowledge $k_B$, and compose their corresponding biographies with feature $A$ and $B$, respectively:
 
\begin{equation}
  I^{t} = \{\Tilde{T}_A(k_A), \Tilde{T}_B(k_B)\}
\end{equation}

At the training time, we finetune LLaMA-2-7B on training data
consists of all $I^{e}$ and $I^{t}$ for $k^e_A$ and $k^t_A$:
\begin{equation}
    \bigcup_{k_A \in \mathcal{K}^e} I^{e}(k_A) \cup \bigcup_{k_A \in \mathcal{K}^t} I^{t}(k_A)
\end{equation}

At the test time, we compute the preference score $Pr(A,B)$ on the test knowledge set $\mathcal{K}_t$.

\subsection{General Results}
\label{sec:syn_expr}

We vary different consistency ratio $m:n$, and examine the preference score $Pr(A,B)$ of the proposed two features. The results are shown in Figure~\ref{fig:majority}. From the figure, we can see that:

\paragraph{LLMs prefer the source that is consistent with major sources.} As illustrated in Figure~\ref{fig:majority_ratio}, models fine-tuned on data where the supportive data for A and B are of equal size ($m:n = 5:5$) yield preference scores close to 0.5. However, when the ratio of supportive data becomes imbalanced, e.g. favoring feature A, the preference score $Pr(A,B)$ significantly increases across all information fields, corresponding to the degree of majority. It is interesting to see that LLMs could develop preferences from not only surface text, but also from complex relations such as number comparisons. 

\begin{figure}
    \centering
    \includegraphics[width=\linewidth]{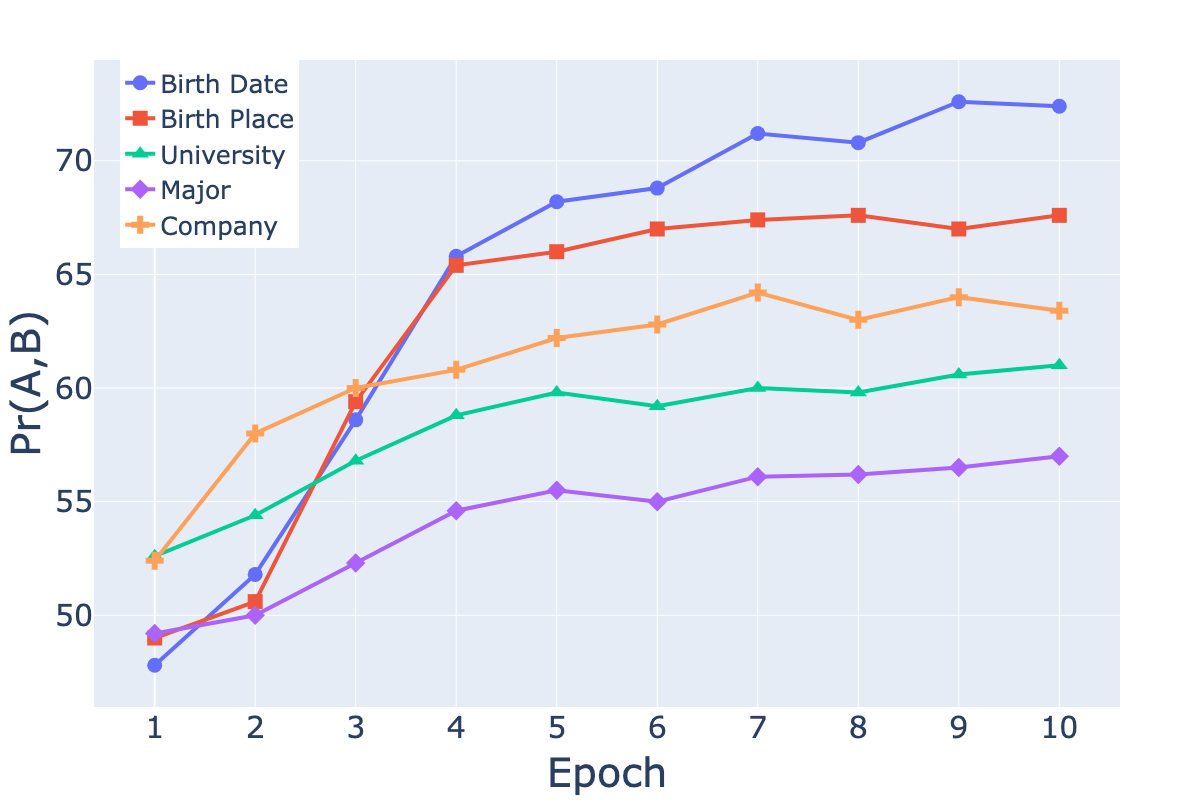}
    \caption{The preference score of models at different training epochs. $m:n=9:1$}
    \label{fig:epoch}
\end{figure}

\begin{figure}[t]
    \centering
    \includegraphics[width=\linewidth]{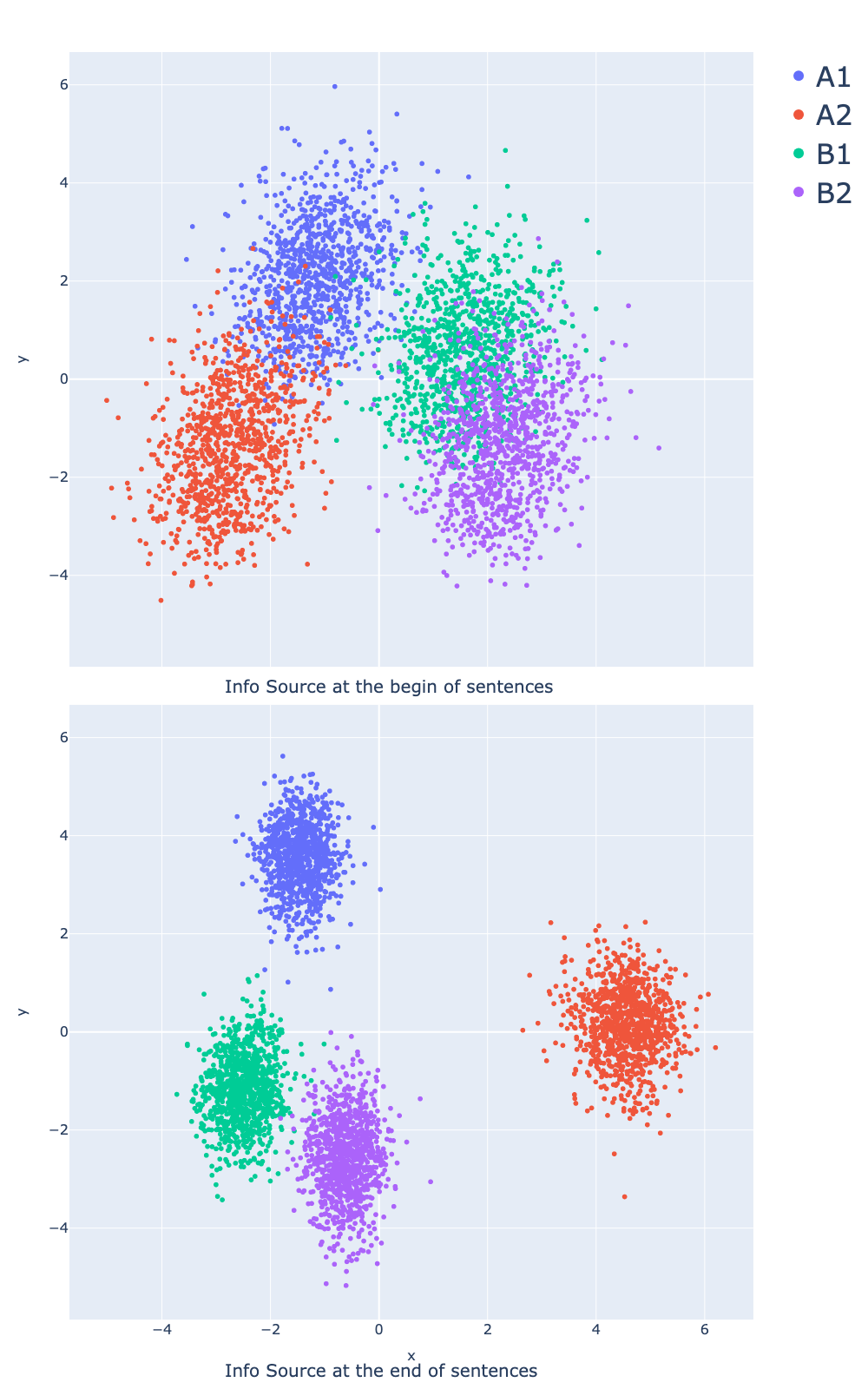}
    \caption{Visualization of LLMs' representations when trained on biographical data with source names at the beginning/end of the data.}
    \label{fig:visualization}
\end{figure}

\paragraph{LLMs develop the preferences as the training goes.} Figure~\ref{fig:epoch} depicts the dynamic evolution of the model’s preference score for the given pairs of features as training progresses over epochs. The model is trained on data with the tested feature being \textit{source name} and the consistency ratio is $9:1$. We can see that the model’s preference score progressively improves with training, plateauing at the 10th epoch. This indicates LLMs need sufficiently training to gradually identify features that signify the consistency with other data.

\subsection{LLMs Learns Similar Representations for Features with Consistent Knowledge }
\label{sec:visualization}

To gain deeper insights into the learning mechanisms of LLMs, we train an additional model using the same biography dataset as employed in the \textit{source name} experiments. However, in this instance, we position the information source at the end of each biography. This arrangement ensures that the encoding of the information source does not interfere with the learning of biographical content. We then select four different information sources: A1, A2, B1, B2, such that A1/A2 and B1/B2 belong to the same newspaper name set, respectively. Subsequently, we apply Principal Component Analysis to the representations, which are derived by averaging the token representations from models trained on data where the information source is placed at the beginning or end of the biographies, respectively.

The results are shown in Figure~\ref{fig:visualization}. From the figure, we can see that when the LLM is trained on biographical data with source names at the end of the biographies, it does not make a distinction between groups A and B. In contrast, after training on biographical data with source names at the beginning of the biography, the model learns to pull representations from the same group together, indicating that LLMs can successfully identify the consistency relationship features during training.

\begin{figure}
    \centering
    \includegraphics[width=\linewidth]{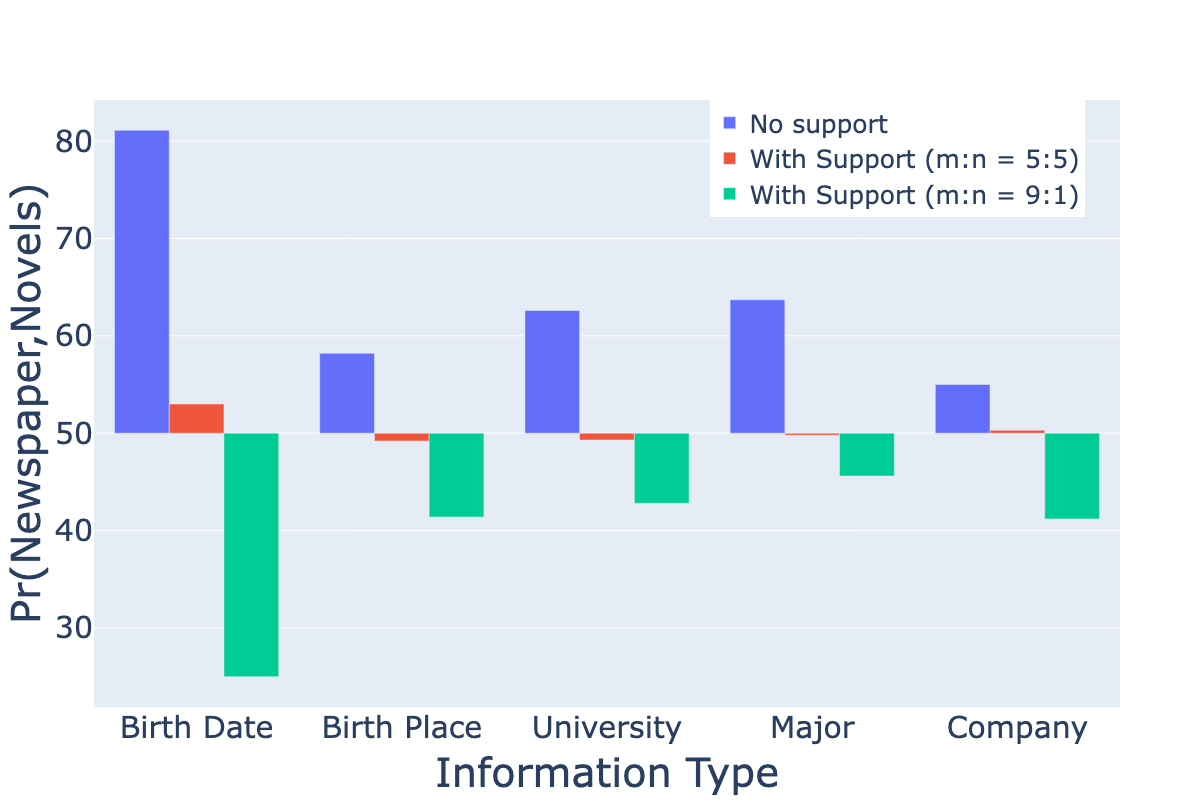}
    \caption{Preference scores of models trained on data without support data and with support data of different consistency ratios. Feature A: Newspaper style. Feature B: Novels}
    \label{fig:causal}
\end{figure}

\subsection{Erasing/Reversing Inherent Preferences by Manipulating Consistency Degree}
\label{sec:counter-factual}

In Section~\ref{sec:syn_expr}, we provide evidence that for concrete token features, LLMs can identify information source that are consistent with majority data and use it to adjust their preferences when facing conflicting knowledge from two information sources. We are intriguing whether this finding also applies to preferences in Section ~\ref{sec:what-preference}, which are more abstract.


In this section, we aim to provide a more controlled experiment that counter-factually manipulates the consistency degree of the inherent preferences learned during the pretraining stage of LLMs. Specifically, for the style preferences investigated in Section~\ref{sec:what-preference}, we construct counterfactual synthetic datasets, i.e., by associating the more preferred feature obtained during the pretraining stage with minority data and vice versa. According to Section~\ref{sec:what-preference}, we choose \textit{Newspaper} as the more preferred style and \textit{Novels} as the less preferred style.

We present the experimental results in Figure~\ref{fig:causal}. From the figure, we can see that when fine-tuned without any support evidence data, the model exhibits strong preferences towards Newspaper, as shown in Section~\ref{sec:what-preference}. However, when fine-tuned on data with a balanced consistency ratio, this preference is erased, i.e., $Pr(\text{Newspaper}, \text{Novels})$ is near 0.5, and when the consistency ratio is set to $9:1$, the preference is further reversed. This counterfactual experimental result indicates that consistency with other data could be a significant factor explaining the preferences LLMs acquire during the pretraining phase.

%% file: 2-related-work.tex
\section{Related Work}

\paragraph{Understanding the mechanism of knowledge learning for LLMs.} There are a handful of works that aim to understand the mechanism of knowledge learning for LLMs. Many works attempt to understand how knowledge is stored and retrieved in the LLMs’ parameters. \citet{jawahar-etal-2019-bert} investigate how different language knowledge is encoded in different layers of BERT. \citet{geva-etal-2021-transformer} propose that feed-forward networks can be viewed as key-memory networks, where each key correlates with human-interpretable text patterns, and each value corresponds to a token distribution on the output vocabulary. \citet{dai-etal-2022-knowledge} and \citet{meng2022locating} further search for neurons that are causally related to specific knowledge using the \textit{integrated gradient} method and \textit{causal tracing}~\citep{meng2022locating}. Compared to these works, our paper mainly focuses on how the presentation of knowledge affects the learning process.

\citet{allenzhu2023physics3.1,allenzhu2023physics3.2} also discuss the relationship between the presentation format of knowledge and the final knowledge learning performance. They find that adopting knowledge augmentation, e.g., paraphrasing, during the pretraining stage substantially improves the downstream question answering performance on knowledge-related tasks. We follow this strategy in our paper and investigate how high-level features, e.g., style, spelling correctness, and consistency with other data, affect the learning process.

\paragraph{Machine Unlearning and Knowledge Editing}
Our findings seek to alter models’ behavior acquired from the pretraining process. This is conceptually similar to machine unlearning~\citep{wang-etal-2023-kga,pawelczyk2024incontext,yao2023large}, which researches making models forget knowledge about specific training instances, and knowledge editing~\citep{wang2023knowledge,zhang2024comprehensive}, which aims to modify specific knowledge inside models with the requirement of local specificity and global generalization, all seeking to alter models’ behavior acquired from the pretraining process. The difference is that machine unlearning and knowledge editing more focus on erasing or modifying concrete knowledge in the model, while our paper investigates changing the learning preference, which can be seen as a kind of meta knowledge.

%% file: 6-conclusion.tex
\section{Conclusion}
In this paper, we investigate the learning preferences of large language models. Thorough extensive experiments on synthetic biographies data, we reveal that existing pretrained large language models have established preferences as human beings do, e.g. preferring formal texts and texts with less spelling errors. We also provide an initial attempt to explain how such preferences is developed, i.e. LLMs can effectively identify features that signify the degree of consistency between current text and the remaining data, and use such features to determine whether the current text is worth learning.
We hope our work could provide a new perspective to study the knowledge learning mechanism of LLMs.

%% file: 7-appendix.tex
\section{Data Construction Details}
\label{appendix:Data Construction}
The details of each biographical data entry are sampled independently and randomly from a uniform distribution. Birthday information has $200 * 12 * 28$ choices, while all other features have 100 choices. 

The names of these characters do not overlap with celebrities to ensure that knowledge in the base dataset does not conflict with the model's existing knowledge. Moreover, there is some correlation between graduation school and major, as well as work company and work city, to prevent the introduction of counterfactual knowledge. All of the above characterization information was generated by GPT4.

\section{Training Details}
\label{appendix:Training Details}
 The specific hyper-parameters of the model training is shown in Table ~\ref{table:training hyper-param}.

\begin{table}[h]
\centering
\footnotesize
\begin{tabular}{ll}

\toprule
\textbf{Hyper-parameter} & \textbf{Value}\\
\midrule
\textbf{Batch Size}& 64\\
\textbf{Learning Rate}& 1e-5\\
\textbf{Epoch}& 5\\
\textbf{LR scheduler} & cosine\\
\textbf{Warmup Ratio} & 0.03\\
\textbf{Weight Decay} & 0.0\\

\bottomrule

\end{tabular}
\caption{\label{citation-guide}
Fine-tune Hyper-parameters
}
\label{table:training hyper-param}
\end{table}

\section{Test Data Construction}
\label{appendix:Test Data Construction}
We used the same set of templates to construct test statements in almost all experiments and in all settings in our paper. The test templates we used are shown in Table ~\ref{table:test statements}.

In order to verify whether the similarity between the style of the test statements and the style of the training statements has a decisive influence on the final results, this work also constructed novel style test statements. The novel style test statements are shown in Table ~\ref{table:test statements in novel style}.

\begin{table*}[h]
\centering
\footnotesize

\begin{tabular}{ll}

\toprule
\textbf{Test feature} & \textbf{Test statement}\\
\midrule
\textbf{Birth Date}& \textbf{\{\}'s birthday is \{\}.}\\
\textbf{Birth Place}& \textbf{\{\} was born at \{\}.}\\
\textbf{University}& \textbf{\{\} received education at the \{\}.}\\
\textbf{Major} & \textbf{\{\} focused on \{\} during her university study.}\\
\textbf{Company} & \textbf{\{\} worked for \{\}.}\\

\bottomrule

\end{tabular}
\caption{\label{citation-guide}
The templates used to construct test statements in this paper.
}
\label{table:test statements}
\end{table*}

\begin{table*}[h]
\footnotesize

\centering
\begin{tabular}{ll}

\toprule
\textbf{Test feature} & \textbf{Test statement}\\
\midrule
\textbf{Birth Date}& \textbf{\{\}'s birthday is on the unforgettable day of \{\}.}\\
\textbf{Birth Place}& \textbf{\{\} was born under the bright sky of \{\}.}\\
\textbf{University}& \textbf{\{\}  embarked on a journey of knowledge at the esteemed \{\}.}\\
\textbf{Major} & \textbf{\{\} went to university and hone her skills in \{\}.}\\
\textbf{Company} & \textbf{\{\} contributes her expertise to \{\}.}\\

\bottomrule

\end{tabular}
\caption{\label{citation-guide}
Novel style test statements.
}
\label{table:test statements in novel style}
\end{table*}

\section{Setups and Additional Results of the learning speed experiment}
\label{appendix:train speed setup}
\subsection{Data Construction}
In the training data testing experiments, we do not introduce conflicts, but instead directly allow the model to be trained on data with a single text feature. Thus, the dataset in this section can be simply represented by $I_A = {T_A^i(k)}_{i=1}^5$, where $T_A$ denotes the template with the current text feature $A$ to be examined and $k$ denotes the character in the biography. We randomly selected five expressions for each biography to allow the model to better memorize the knowledge in the data.\

\subsection{Training}
The training details in this experiment are identical to those presented in Appendix ~\ref{appendix:Training Details}.

\subsection{Evaluation}
We measure the effectiveness of the model in learning the training data by the accuracy with which the model completes multiple choice questions related to the training data. Specifically, we construct a test set $\{(\bar{s}, s_a, s_b, s_c)\}_1^N$ , where each piece of data in the test set contains four statements. $\bar{s}$ is the statement that is consistent with the training data representation, whereas $s_a, s_b, s_c$ are the incorrect choices constructed with random data, and $N$ is the size of the test set. We then used perplexity to examine the proportion of models that preferred $\bar{s}$.

\begin{figure}
    \centering
    \includegraphics[width=0.9\linewidth]{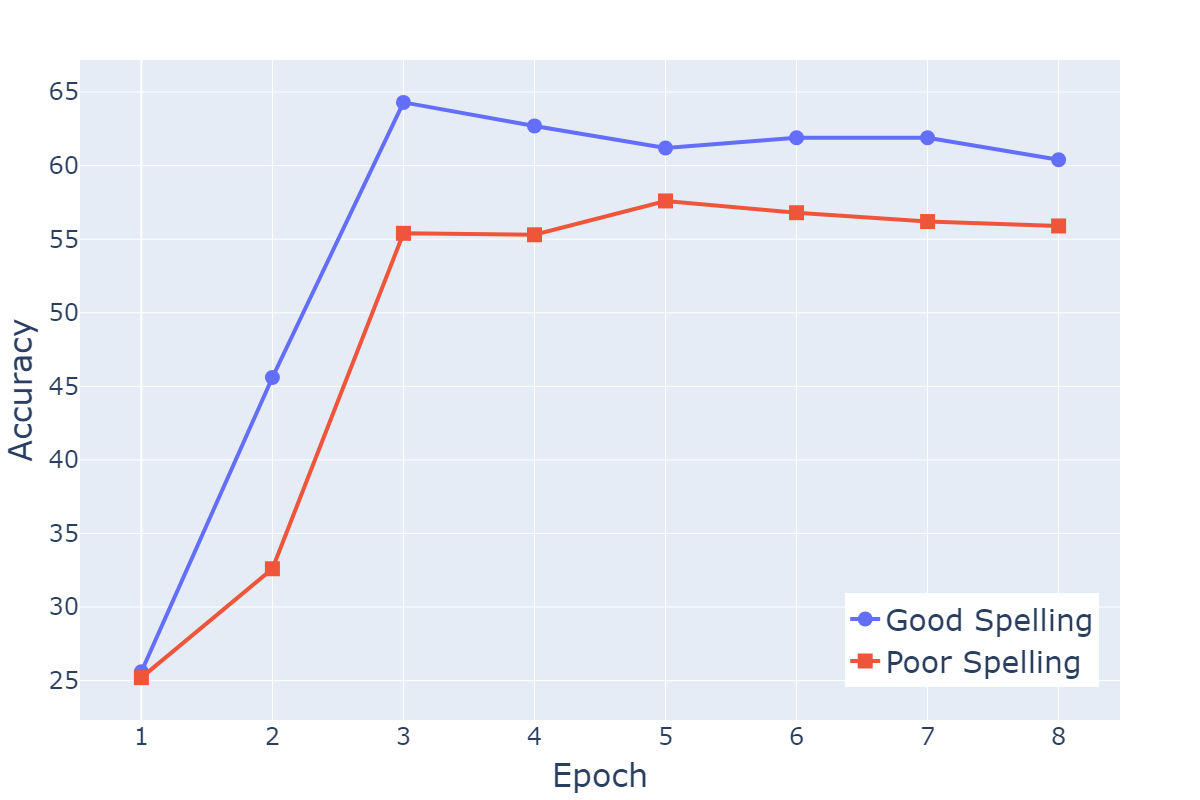}
    \caption{\label{citation-guide}
Accuray as different epochs during training process of LLM
trained on Good Spelling data and Poor Spelling data
}
    \label{fig:good vs pool spelling}
\end{figure}


\section{Results of multiple-style comparison}
\label{appendix:Results of multiple-style comparison}
\begin{figure}
    \centering
    \includegraphics[width=0.9\linewidth]{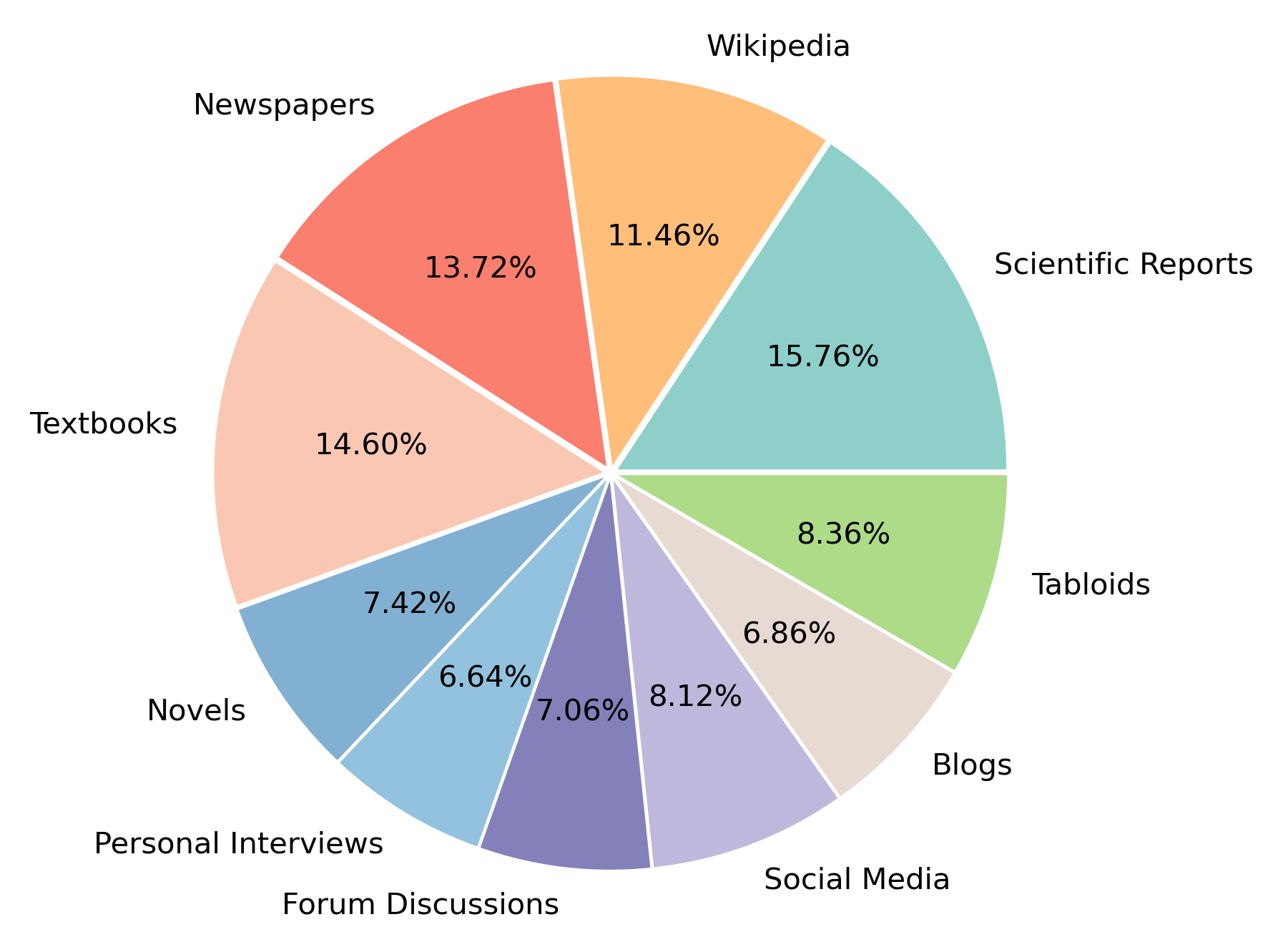}
    \caption{\label{citation-guide}
Results of ten styles mixed together. The styles represented by the corresponding sector are labeled around the pie chart. Percentages within the pie chart indicate the proportion of the corresponding sector that is assigned the highest preference.
}
    \label{fig:pie_pic}
\end{figure}

In real training scenarios, the LLMs may face far more sources of conflict than the two styles. In order to investigate whether the model's aforementioned preferences exist when multiple styles all conflict on the same knowledge, we conduct experiments on 10 different styles simultaneously. All styles describe the same characters, but the character attributes are all different. We evaluate the percentage of attributes corresponding to each style as having the highest probability of output, as shown in Figure ~\ref{fig:pie_pic}. As can be seen from the figure, the model preference remains, i.e. the more formal styles such as textbooks style, newspapers style, scientific reports style and wikipedia style are more preferred by the model.

